\documentclass[conference]{IEEEtran}
\IEEEoverridecommandlockouts
\usepackage{amsmath,amssymb,amsfonts}
\usepackage{algorithmic}
\usepackage{graphicx}
\usepackage{textcomp}
\usepackage{siunitx}
\usepackage{booktabs}
\usepackage{xcolor}
\usepackage{authblk}

\usepackage[sorting=none]{biblatex}
\def\BibTeX{{\rm B\kern-.05em{\sc i\kern-.025em b}\kern-.08em
    T\kern-.1667em\lower.7ex\hbox{E}\kern-.125emX}}
\bibliography{lit.bib}
\begin{document}
\title{Tumor likelihood estimation on MRI prostate data by utilizing k-Space information}
\author[1, 2, 5]{Moritz Rempe\thanks{\hspace*{-1.1em}Corresponding author: moritz.rempe@uk-essen.de}\thanks{This work received funding from the Bruno \& Helene Jöster Foundation, enFaced (FWF KLI 678), enFaced 2.0 (FWF KLI 1044) and KITE (Plattform für KI-Translation Essen) from the REACT-EU initiative (https://kite.ikim.nrw/).
The authors declare no competing interests.}}
\author[1, 2, 5]{Fabian Hörst}
\author[1]{Constantin Seibold}
\author[4]{Boris Hadaschik}
\author[5]{\authorcr Marco Schlimbach}
\author[1, 2, 6, 7, 8]{Jan Egger}
\author[5]{Kevin Kröninger}
\author[3]{Felix Breuer}
\author[3]{Martin Blaimer}
\author[1, 5, 9]{Jens Kleesiek}
\affil[1]{Institute for AI in Medicine (IKIM), University Hospital Essen, Girardetstraße 2, 45131 Essen, Germany}
\affil[2]{Cancer Research Center Cologne Essen (CCCE), University Medicine Essen, Hufelandstraße 55, 45147 Essen, Germany}
\affil[3]{Magnetic Resonance and X-ray Imaging Department (MRB), Fraunhofer Institute for Integrated Circuits IIS,\protect\\ Am Hubland, 97074 Würzburg, Germany}
\affil[4]{Department of Urology, University Hospital Essen, Hufelandstraße 55, 45147 Essen, Germany}
\affil[5]{Department of Physics of the Technical University Dortmund, Otto-Hahn-Straße 4a, 44227 Dortmund, Germany}
\affil[6]{Computer Algorithms for Medicine Laboratory, 8010 Graz, Austria}
\affil[7]{Institute of Computer Graphics and Vision, Graz, Austria}
\affil[8]{University of Technology, Inffeldgasse 16, 8010 Graz, Austria}
\affil[9]{German Cancer Consortium (DKTK), Partner Site Essen, Hufelandstraße 55, 45147 Essen}
\setcounter{Maxaffil}{0}
\renewcommand\Affilfont{\itshape\small}

\maketitle

\begin{abstract}

We present a novel preprocessing and prediction pipeline for the classification of magnetic resonance imaging (MRI) that takes advantage of the information rich complex valued k-Space. Using a publicly available MRI raw dataset with 312 subject and a total of 9508 slices, we show the advantage of utilizing the k-Space for better prostate cancer likelihood estimation in comparison to just using the magnitudinal information in the image domain, with an AUROC of 86.1\%±1.8\%. Additionally, by using high undersampling rates and a simple principal component analysis (PCA) for coil compression, we reduce the time needed for reconstruction by avoiding the time intensive GRAPPA reconstruction algorithm. By using digital undersampling for our experiments, we show that scanning and reconstruction time could be reduced. Even with an undersampling factor of 16, our approach achieves meaningful results, with an AUROC of 71.4\%±2.9\%, using the PCA coil combination and taking into account the k-Space information. With this study, we were able to show the feasibility of preserving phase and k-Space information, with consistent results. Besides preserving valuable information for further diagnostics, this approach can work without the time intensive ADC and reconstruction calculations, greatly reducing the post processing, as well as potential scanning time, increasing patient comfort and allowing a close to real-time prediction.

\end{abstract}


\section{Introduction}
\label{sec:introduction}
\IEEEPARstart{M}{agnetic} Resonance Imaging (MRI) is often the favored medical imaging technique for clinical examination due to its high resolution, soft tissue contrast and lack of harmful exposure to radiation. One drawback of MRI acquisitions is the long imaging time, which can not only lead to artifacts in the image itself due to patient movement, but also be exhausting for the patient. To speed up the imaging process, nowadays, scanner acquire the image information with multiple coils and often use a technique called \textit{undersampling} \cite{heidemann2003brief}\cite{blaimer2004smash}. The concept of undersampling involves acquiring only a reduced amount of information and subsequently reconstructing it, resulting in imaging acceleration of up to 48 times \cite{Belov_Stadelmann_Kastryulin_Dylov_2021}, although in clinical practice, it is often limited to up to 10 times.

Even though the scanning time for the patient is greatly improved, the undersampled data leads to wrapping artifacts as well as lower signal values, decreasing the signal-to-noise ratio (SNR) by by the square root of the undersampling rate. As a consequence, complex reconstruction methods, such as for example the commonly used algorithms \textit{GRAPPA} \cite{Griswold_Jakob_Heidemann_Nittka_Jellus_Wang_Kiefer_Haase_2002} or \textit{SENSE} \cite{pruessmann1999sense} have to be applied on the raw data, which can be very time intensive and introduce additonal noise amplification (also known as geometry factor, g-factor) \cite{breuer2009general}.

With a fatality rate of 11\% and being the most frequent malignancy among men in 2023 \cite{Siegel_Miller_Wagle_Jemal_2023}, the detection for prostate cancer (PCa) is an urgent task.
Because downstream tasks, such as segmentation and classification rely on fully reconstructed image data, there is a great urge to reduce the reconstruction time. Due to advances in the field of deep learning in the past years, the quality of reconstruction of the undersampled data has greatly improved, while reducing the computation times \cite{Pal_Rathi_2022}\cite{Hyun_Kim_Lee_Lee_Seo_2018}. 
While all common reconstruction algorithms output an image interpretable for radiologist and other clinicians,  raw MRI data is acquired in the so called k-Space. This complex valued raw data, which is normally discarded after reconstruction, holds more information than currently used by downstream tasks \cite{Paschal_Morris_2004}, such as tumor classification. 

In this work, we show that by utilizing the MRI k-Space, classification on MRI data can be improved. Additionally, due to the elimination of time-intensive reconstruction techniques, the data acquisition and processing time can be greatly reduced. With a highly reduced time needed for the initial prediction, our approach can be used for real-time diagnostics to conduct further examinations right away, without greatly impacting patient comfort.

We investigate the benefits of taking into account all available raw data information in MRI scans on the performance of prostate cancer prediction. The prediction will be based on the probability given by the Prostate Imaging Reporting and Data System (Pi-RADS) scores above a score of two. We also compare the approach of PCA coil combination with the more time consuming reconstruction technique GRAPPA. 

\begin{figure*}[!t]
  \centering
  \includegraphics[width=\textwidth]{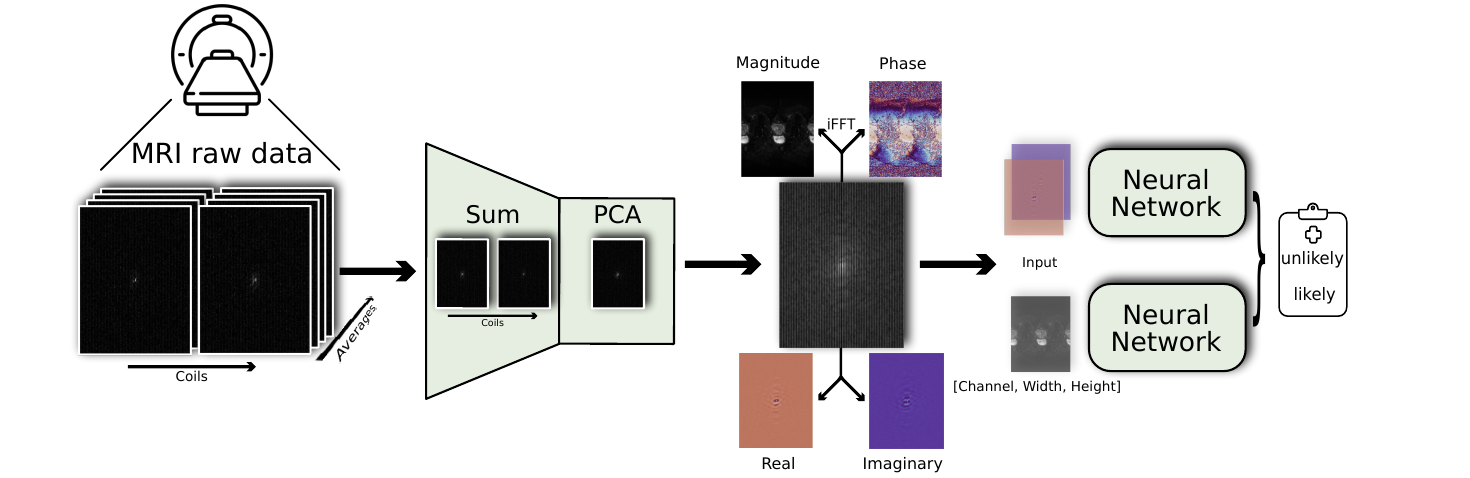}
  \caption{Methodical overview of the classification pipeline via PCA. The complex valued MRI raw data 2D slices are summed over the averages and then fed into the PCA. The first PCA component is then split into its image domain magnitude and phase, as well as its k-Space real and imaginary part. With this stacked input, the ConvNeXts then predict if the slice should be classified with a Pi-RADS score larger than two.}
  \label{fig:method_overview}
\end{figure*}

The code used for this project is publicly available at \url{https://github.com/TIO-IKIM/tumor-prediction-on-undersampled-MRI-kSpace}.

\section{Related Work}

For prostate cancer detection and classification, multiple different clinical methods can be used, such as \cite{cornford2024eau}:
\begin{itemize}
 \item prostate-specific antigen (PSA),
 \item digital rectal exam,
 \item trans-rectal ultrasound (TRUS),
 \item multiparametric magnetic resonance imaging (mpMRI).
\end{itemize}
Because the first three methods often lead to more discomfort for the patient, today MRI, especially multiparametric MRI, including diffusion weighted imaging (DWI), is the preferred method for detecting PCa \cite{srivastava2024use}\cite{asif2023comparing}.
The scans are commonly ranked by the clinical Pi-RADS score in a range between one (clinically significant cancer is highly unlikely) to five (clinically significant cancer is highly likely) \cite{turkbey2019prostate}.

One major issue of MRI is the aspect of time needed for performing the scan itself, with DWI scans taking up to 10 minutes. By reducing the imaging time, which also increases the comfort for the patient, more subjects can be scanned in the clinical routine. One common way of reducing the scanning time is to \textit{undersample} the so called k-Space, e.g. only scanning every forth k-Space line, thus reducing the overall scanning time by the factor of $\times 4$. Another time consuming aspect of MRI is the k-Space reconstruction after the initial scan, such as GRAPPA, which can take more than 10 additional minutes.

With the advent of machine learning, especially in the medical domain \cite{Egger_Gsaxner_Pepe_Pomykala_Jonske_Kurz_Li_Kleesiek_2022}\cite{egger2021deep}, radiologist have new, powerful tools at their hand to support them in their search for clinically significant PCa. 
Convolutional Neural Networks (CNNs), such as the \textit{ConvNeXt} \cite{Liu_Mao_Wu_Feichtenhofer_Darrell_Xie_2022}, show superhuman performance on a wide range of classification tasks and are adaptable to many different fields of work.
Novel algorithms can detect PCa based on the DWI images automatically \cite{Yoo_Gujrathi_Haider_Khalvati_2019}. Schelb et al. \cite{Schelb_Kohl_Radtke_Wiesenfarth_Kickingereder_Bickelhaupt_Kuder_Stenzinger_Hohenfellner_Schlemmer_etal._2019} showed that their U-Net trained with T2-weighted and diffusion MRI was able to achieve similar performance to Pi-RADS assessment.

But all state-of-the art algorithms for prostate cancer detection have two things in common: they only work on fully sampled or reconstructed data and they perform the detection in the image domain, after an inverse Fourier transformation has been performed and the raw data has been filled.

By performing all these preprocessing steps, including k-Space reconstruction, inverse Fourier Transformation and discard of the phase information, valuable information in the raw data is lost irreversibly.

The \textit{k-Strip} algorithm \cite{Rempe_Mentzel_Pomykala_Haubold_Nensa_Kroeninger_Egger_Kleesiek_2024} showed, that segmentation tasks, such as skull stripping, are possible directly in the MRI raw data domain. The authors implemented a complex valued convolutional neural network, which takes the k-Space as a direct input. In this work we build on the idea to use MRI raw data. We present superior results in detecting likely PCa on DWI raw data, optionally without the need of time expensive reconstruction techniques, such as GRAPPA, allowing a greatly reduced scanning time, while still achieving meaningful predictions.

\section{Material and Methods} 

\subsection{Fourier Transformation}

The Fourier Transformation \cite{bracewell1986fourier} is one of the most important equations in signal theory. It converts a 2D-signal from the time-domain into the frequency-domain or from the spatial-domain into the spatial-frequency-domain.
By using the inverse form, one can convert the raw MRI data (k-Space) into the image domain. This process is completely reversible and does not lose any information, as long as both magnitude and phase are preserved.

\subsection{k-Space \& Data Acquisition}

The so-called k-Space is the complex valued raw data domain of MRIs. By applying the inverse Fourier Transformation, one can convert the data into the complex valued image domain. 
Today, MRI scanner use multiple coils to acquire different parts of the image simultaneously, which are then combined by performing a root square sum (RSS) calculation over all coils in the image domain. By using the sensitivity of each coil, the final image can be modulated out of the resulting set of k-Space matrices. 

Commonly used images for downstream tasks, such as tumor classification, are only the magnitudinal data of the image domain, leading to a loss of phase information. Phase information was shown to be useful for many downstream tasks, an example being flow detection \cite{Wymer_Patel_Burke_Bhatia_2020}. 

\subsection{Datasets}
The amount of publicly available datasets which include MRI raw data is scarce, as this data is not commonly saved in clinical practice. We are only aware of a single dataset, suitable for classification tasks. Nevertheless we are able to show the advantage of using the raw data over only using magnitudinal data. 

\begin{figure}[!t]
  \includegraphics[width=\columnwidth]{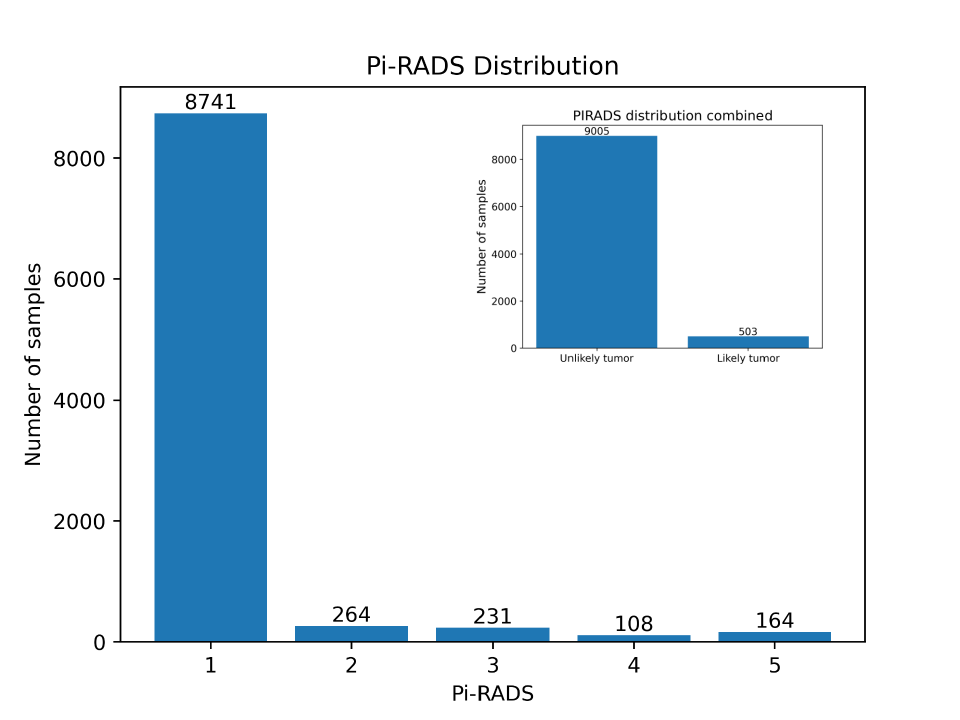}
  \caption{Pi-RADS label distribution of the FastMRI Prostate dataset, as well as the distribution after combining the classes.}
  \label{fig:data_distribution}
\end{figure}

The FastMRI Prostate dataset \cite{Tibrewala_Dutt_Tong_Ginocchio_Keerthivasan_Baete_Chopra_Lui_Sodickson_Chandarana_etal._2023} contains T2 and DWI raw data with different amounts of coils (14 - 30) and slices (24 - 38), with a in-plane resolution of $2.0\times2.0$ \si{\milli\meter}. 
The data was acquired on two clinical 3\si{\tesla} systems (MAGNETOM Vida, Siemens Healthineers, Germany).
The DWI raw data has a matrix size of $100\times100$ and a resulting field of view (FOV) of $200\times200$ \si{\milli\meter}.
The diffusion consists of three directions and four averages for the B50 images, three directions and 12 averages for the B1000 images and two averages for the B0 images. B1500 images and apparent diffusion coefficients (ADC) maps have to be calculated after reconstruction and are not directly present in the raw data. 
In total, this dataset contains 312 male patients with a total of 9508 slices with corresponding Pi-RADS scores, reviewed by a radiologist. The majority of subjects in the dataset had elevated PSA values and underwent pre-biopsy imaging. The Pi-RADS classification has been performed on the fully reconstructed magnitudinal data only. The dataset comes with k-Space data of both T2 and DWI. To show the feasibility of our approach, we only work on the DWI data, as it needs more extensive post-processing steps. That way we can show, that even when multiple postprocessing steps are necessary for the image domain, the raw data can still deliver more valuable information. The authors of the FastMRI Prostate dataset give a predefined 70\% - 15\% - 15\% datasplit for training, validation and test dataset, which was also used in this work. The label distribution of the total dataset can be seen in Fig. \ref{fig:data_distribution}. The resulting splits have a distribution similar to the total label distribution.

\begin{figure*}[!htb]
  \centering
  \includegraphics[width=\textwidth]{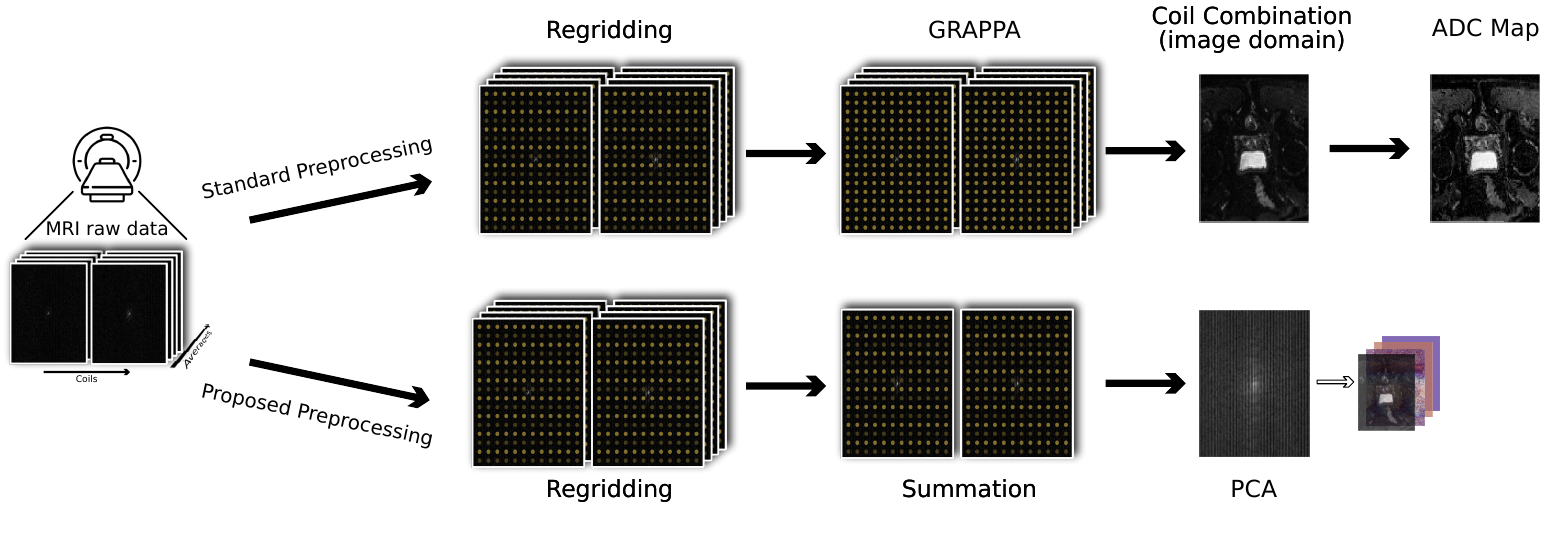}
  \caption{Comparion of the standard preprocessing pipeline and the proposed preprocessing pipeline. The standard pipeline includes regridding, GRAPPA reconstruction, followed by coil combination in the image domain and the adc map calculation. The phase information are lost. The proposed pipeline first performs regridding, followed by a summation and coil compression via PCA. This approach is less computational intensive and preserves the complex valued k-Space data.}
  \label{fig:preprocessing}
\end{figure*}

Because of the high imbalance of the dataset, the tumor prediction is performed as a binary problem, with all scans having a Pi-RADS score of $\leq 2$ belonging to class 0 (unlikely tumor) and all slices with a score $>2$ belonging to class 1 (further examination in exceptional case up to highly likely tumor). These thresholds have been chosen according to the original dataset publication.

\subsection{Preprocessing \& Network Architecture}
To get from the initial MRI raw data to the final image used in clinical workflows, a lot of preprocessing steps, some being very time intensive, have to be performed. In the case of the DWI raw data, the shape of the k-Space is [Average, Coil, Depth, Width, Height]. The averages are used to calculate the final data over multiple scans to reduce noise. Each coil in the data covers a different part of the image with different intensity. 

The first necessary preprocessing steps are calculating the data average and combining the coils. A comparison of the standard preprocessing pipeline and the proposed pipeline is showed in Fig. \ref{fig:preprocessing}.

To simplify the averaging calculation, instead of splitting the DWI data into b50, b1000 and adc channel, which will be subject of future work, we sum all averages into one channel. In the next step, the data is then either fed into the principal component analysis (PCA) \cite{Jolliffe_Cadima_2016} pipeline or the GRAPPA algorithm. The PCA transforms the data such that it results in multiple principal components, with a descending amount of variance in the data. The very first principal component is the k-Space of interest for us. 
While MRI coils are typically combined in the image domain by using their coil sensitivity maps or RSS, PCA can compress the coils directly in the k-Space.
Because PCA is a compression method, information are lost when just using the very first principal component. Nevertheless, in our experiments it showed, that in the case of DWI data, using more than the first component does not attribute to better classification results. 
The resulting data will be the sum over all averages and compressed into one single coil. 

\begin{figure*}[!htb]
  \centering
  \includegraphics[width=15cm]{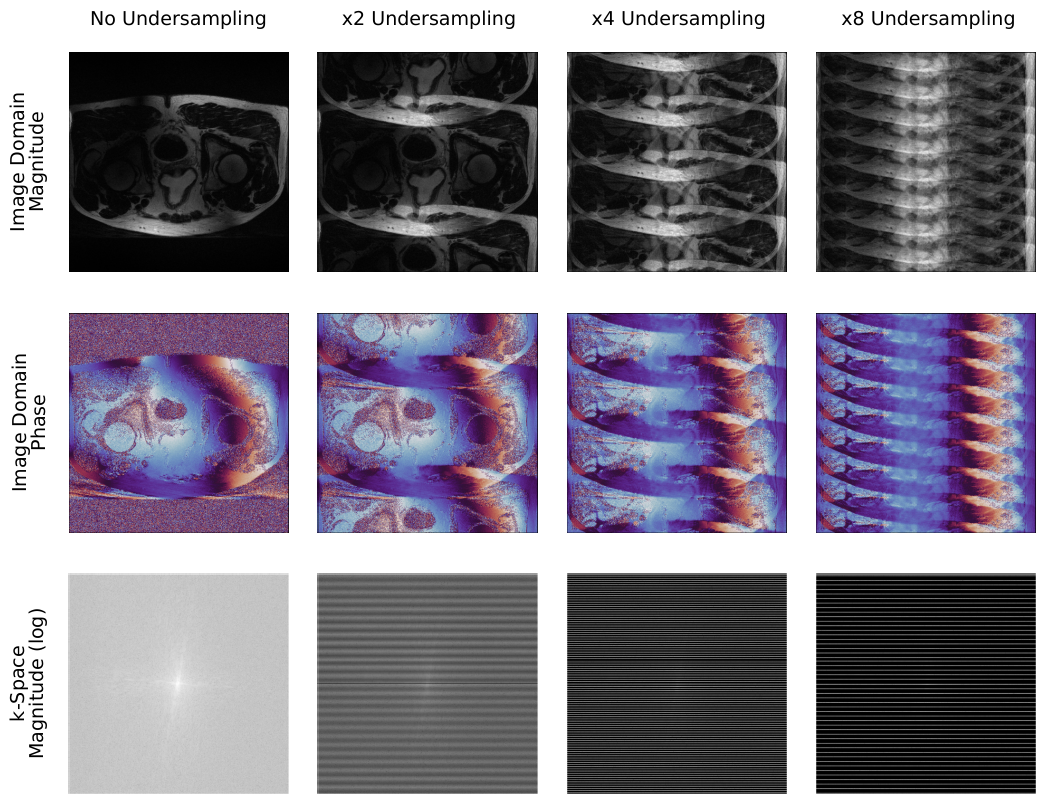}
  \caption{Example of Undersampling in image domain from a fully sampled sample of the FastMRI+ Dataset. Undersampling refers to the missing out of lines in the k-Space. While reducing the imaging time, this leads to artifacts in the image domain, as seen in the top rows. The image wraps in itself, depending on the undersampling factor.}
  \label{fig:undersampling_example}
\end{figure*}

The GRAPPA algorithm interpolates missing multi-coil k-Space signals by convolution.
Coil combination is done after FFT in the image domain, e.g. by a root-sum-of squares combination or an adaptive coil-combination \cite{walsh2000adaptive}. 

The general implementation of the GRAPPA algorithm does not save the k-Space information, as it is supposed to output a reconstructed magnitudinal image in the image domain. We adapted it such that intermediate k-Space results are saved. In our experiments, to preserve the k-Space, we again use a coil compression via PCA after the GRAPPA reconstruction to. Thus, both PCA and our GRAPPA implementation outputs are complex valued and still in the frequency domain. To feed the data into the real valued ConvNeXts, the data is split into magnitude and phase in the image domain, by applying the inverse Fourier transformation, as well as into real and imaginary parts in the k-Space. Depending on the experiment, these parts are then stacked as multiple channels and fed into the networks. 

In case of the GRAPPA and "gold-standard" data, the image domain data and k-Space data are given into two separate neural networks. The output it then averaged and back-propagated through both networks.

Exemplary data can be seen in Fig. \ref{fig:undersampling_example}.

The top row shows the resulting magnitudinal image domain data, the middle row the corresponding phase information. The phase is wrapped into the range of $[-\pi;\pi]$. The bottom row shows the magnitudinal k-Space data, which is plotted logarithmically for better visuality.

It has to be stated, that our implementation is not optimized as it is for algorithms directly implemented on MRI scanner. These implementations are often optimized for the specific vendor and can perform quicker reconstructions. The following experiments and results are supposed to emphasize the speed-up achieved by performing predictions at higher undersampling rates. Using PCA data instead of performing reconstruction can save additional time and also reduce the energy footprint, but the extent of these benefits depends on the final implementation on the scanner.

As a state-of-the-art neural network, we implement the ConvNeXt with roughly 3 million parameters using the \textit{timm} framework \cite{rw2019timm}- This framework allows us to easily change the number of channels of ConvNeXt. The network is pretrained on ImageNet \cite{5206848}.

An overview of the pipeline is shown in Fig. \ref{fig:method_overview}.

\subsection{Undersampling Augmentation}

A key ingredient for the successful training of a neural network is data augmentation. Training data is altered to artificially create and thus increase the available data. By applying multiple augmentations, the network becomes more robust against unseen data. We propose a new undersampling method, tailored for MRI data, the simulated undersampling augmentation.
To artificially undersample the data and to implement the undersampling augmentation, a cartesian undersampling pattern is applied, by evenly cutting out every $n$-th line in the k-Space, as can be seen in Fig. \ref{fig:undersampling_example} (bottom). We only simulate the strategy of evenly undersampling the k-Space, as it is the commonly used technique in DWI. Future experiments will show different undersampling patterns on other imaging sequences.

\section{Experiments}
\label{sec:experiments}

\subsection{Training}

In a first experiment, we compare the prediction results on the "gold-standard" data with and without additionally k-Space information. The image domain data consists of reconstructed data via GRAPPA and combined coils with the coil sensitivity maps. The additional k-Space data is compressed after the GRAPPA reconstruction via PCA and no further postprocessing is performed. The two methods are compared on the fully reconstructed data, as well as at the native undersampling rate of x2. For this experiments, both models are trained with undersampling rates of up to x2.

In the second experiment, we compare the GRAPPA reconstruction method with the PCA pipeline to show the theoretical time benefits at higher undersampling rates of up to x64. The undersampling is digital by removing lines in the k-Space and thus does not fully simulate real undersampling at these high rates. The b50 data is summed over the x, y and z values to create one channel for each coil and then compressed via a PCA. Both models are trained with simulated undersampling rates of up to x8 and tested on rates of up to x64.

For training, the Adam optimizer \cite{kingma2014adam} with an initial learning rate of 1e-4, a beta coefficient of 0.99 and an epsilon value of 1e-08 is used. The learning rate is scheduled using a cosine annealing schedule. Early stopping with a patience of 10 is used to stop the training when the validation loss is not further decreasing. The networks predictions are compared with the ground truth using the Cross-Entropy loss with a weight of 17 to 1, reflecting the imbalance of the dataset. 
In all experiments, the training data is augmented by additionally applying random horizontal flips.
All input data is normalized and standardized batch wise and over each channel separately.

With a batch size of 128, the overall training time takes less than 30 minutes on average on an NVIDIA A6000 GPU with 48GB of graphics memory.

The hyperparameter have been selected by applying a grid search for all input types. 

\begin{figure*}[!htb]
  \centering
  \includegraphics[width=17cm]{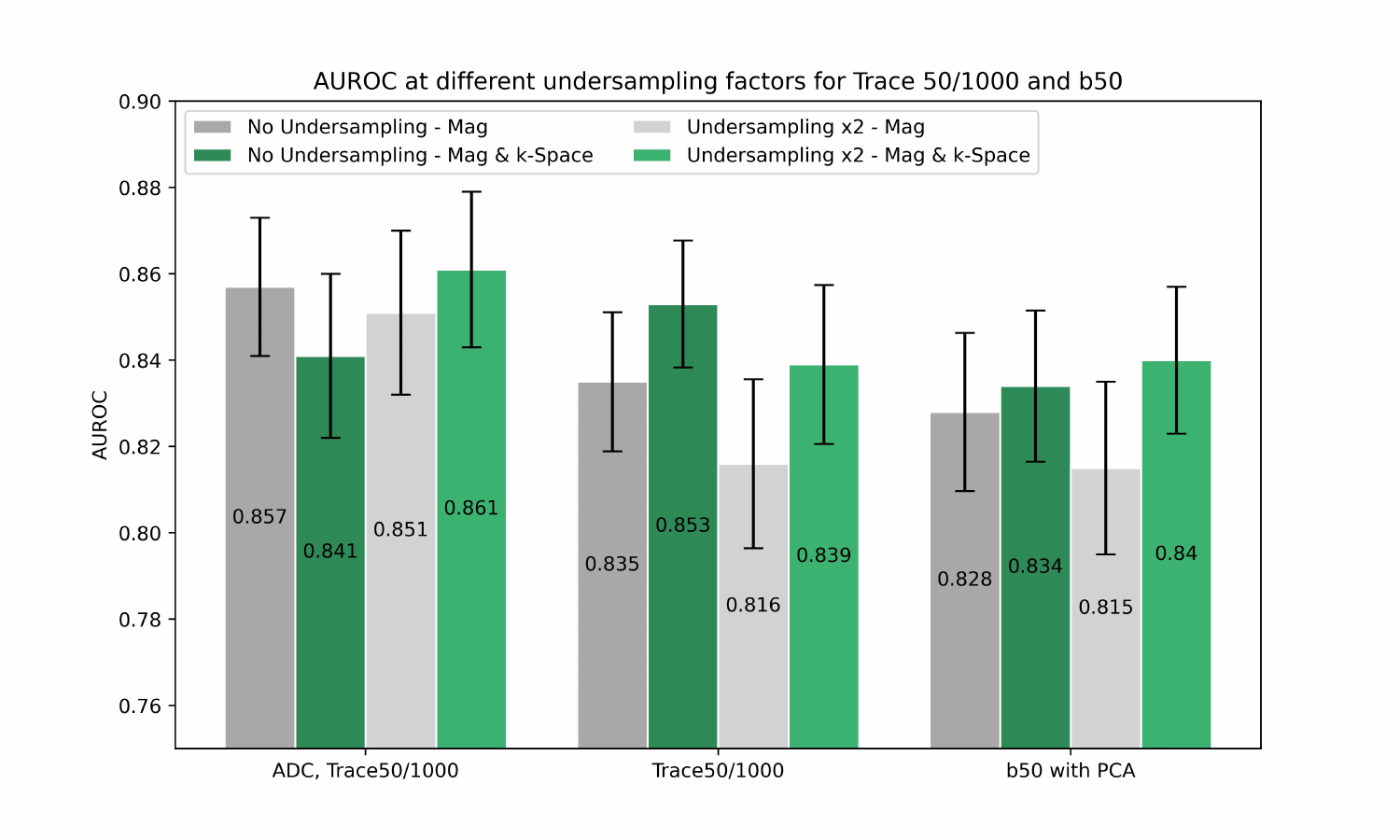}
  \caption{Comparison of the results with and without additional k-Space information on the "gold-standard" data, consisting of GRAPPA reconstructed and coil combined adc, b50 and b1000 maps in the image domain (Mag) with different undersampling factors.}
  \label{fig:x2_results}
\end{figure*}

\subsection{Evaluation Metrics}

For evaluation we decided to use two metrics, one being the area under the receiver operating characteristic curve (AUROC), which describes the area under the curve when plotting the True Positive Rate (TPR) against the False Positive Rate (FPR).

The other metric used is the area under the precision recall curve (AUPRC), describing the area under the curve when plotting the recall against the sensitivity,
\begin{align}
    & \text{Precision} = \frac{\text{TR}}{(\text{TR} + \text{FP})} \\
    & \text{Recall} = \frac{\text{TR}}{(\text{TR} + \text{FN})}
\end{align}
with TR being true positives, FP being false positives and FN being false negatives.

To gather confidence intervals, the bootstrapping method with 1000 iterations has been applied on the test dataset.  

\section{Results}

\begin{table}[!htb]
\caption{Comparison of the results with and without additional k-Space information on the "gold-standard" data, consisting of GRAPPA reconstructed and coil combined adc, b50 and b1000 maps in the image domain with different undersampling factors. x0 means no additional undersampling.}
\centering
\def\arraystretch{1.2}
\resizebox{\columnwidth}{!}{%
\begin{tabular}{lll}
\toprule
\multicolumn{1}{l}{Method \& Undersampling} & \multicolumn{1}{l}{AUROC ($\%$) $\uparrow$} & \multicolumn{1}{l}{AUPRC ($\%$) $\uparrow$} \\
\midrule
Image Domain (ADC, $\text{Trace}_{50}$ \& $\text{Trace}_{1000}$) & & \\
\hspace{1.5cm} x0 & \textbf{85.7±1.6} & \textbf{64.6±2.3} \\ 
\hspace{1.5cm} x2 & 85.1±1.9 & 63.4±2.2 \\
Image Domain \& k-Space & & \\
\hspace{1.5cm} x0 & 84.0±1.9 & 63.5±2.2 \\
\hspace{1.5cm} x2 & \textbf{86.1±1.8} & \textbf{64.3±2.3} \\
\midrule
Image Domain ($\text{Trace}_{50}$ \& $\text{Trace}_{1000}$) & & \\
\hspace{1.5cm} x0 & 83.5±1.6 & 59.5±1.5 \\
\hspace{1.5cm} x2 & 81.6±1.9 & 59.5±1.7 \\
Image Domain \& k-Space & & \\
\hspace{1.5cm} x0 & \textbf{85.3±1.5} & \textbf{62.2±2.0} \\
\hspace{1.5cm} x2 & \textbf{83.9±1.8} & \textbf{60.1±1.7} \\
\midrule
k-Space ($\text{b}_{50}$) & & \\
\hspace{1.5cm} x0 & 82.8±1.8 & 60.5±1.9 \\
\hspace{1.5cm} x2 & 81.5±2.0  & 58.8±1.5 \\
Image Domain \& k-Space & & \\
\hspace{1.5cm} x0 & \textbf{83.4±1.7} & \textbf{63.4±2.3} \\
\hspace{1.5cm} x2 & \textbf{84.0±1.7} & \textbf{62.3±2.2} \\ 
\bottomrule
\end{tabular}}
\label{tab:2}
\end{table}

Fig. \ref{fig:x2_results} shows the results for the fully reconstructed "gold-standard" data. We compare the use of only the image domain data (Mag) with the use of additional k-Space data. The combination of ADC, Trace50 and Trace1000 takes into account all available image domain information. The k-Space data for the ADC map can not be calculated, as it depends on calculations in the magnitudinal image domain. Nevertheless, the model with additional k-Space information achieves better results than its image domain counterpart at an undersampling rate of x2 with an AUROC of $86.1\%\pm1.8\%$. When not taking into account the ADC map, thus eliminating the need for an additional b0 image, the additional k-Space information leads to better results for no additional undersampling ($85.3\%\pm1.3\%$), as well as for x2 undersampling ($83.9\%\pm1.8\%$). Additional AUPRC values can be found in Tab. \ref{tab:2}.

\begin{figure*}[!htb]
  \centering
  \begin{minipage}[b]{0.45\linewidth}
    \centering
    \includegraphics[width=\linewidth]{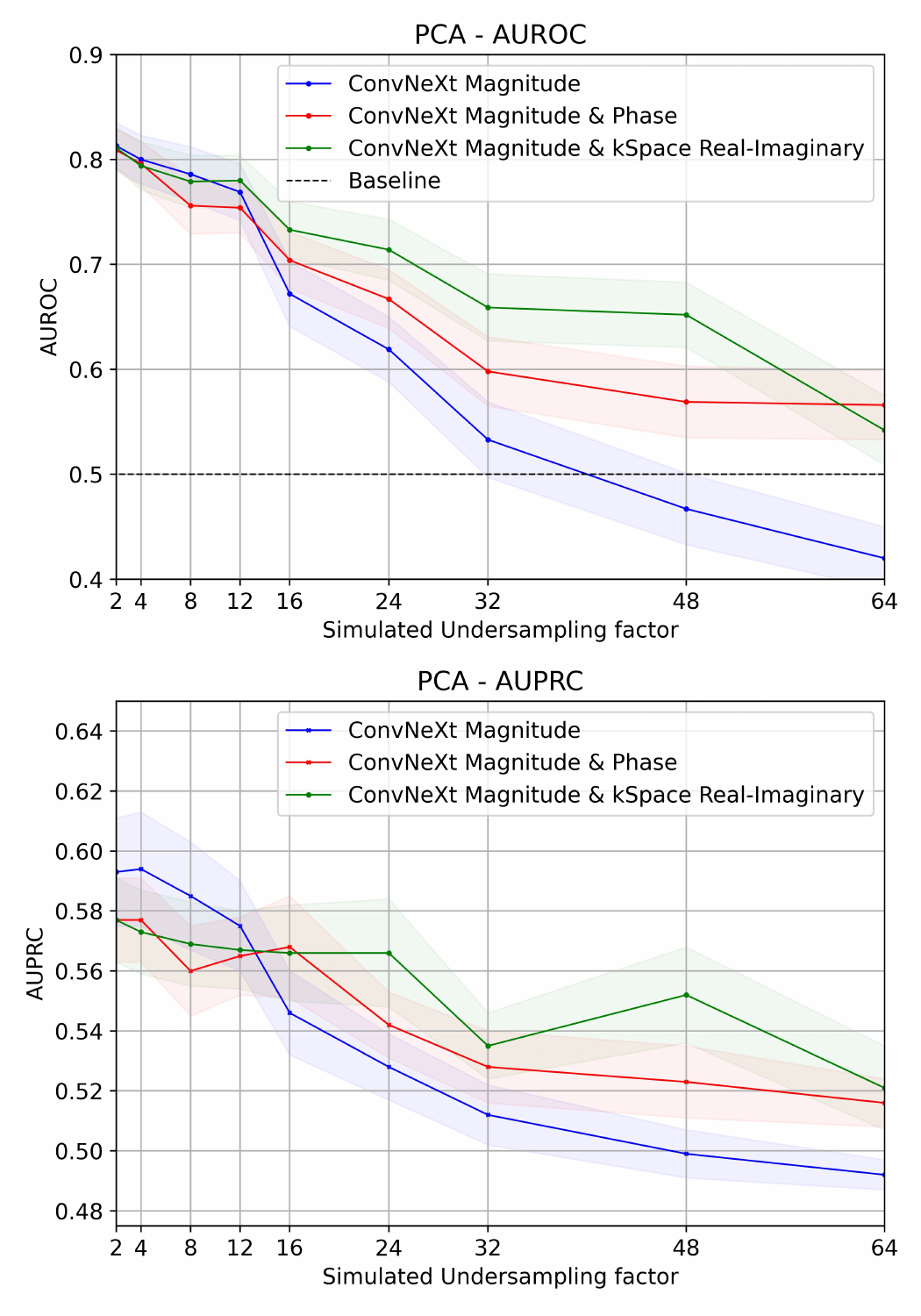}
    \label{fig:results_pca}
  \end{minipage}
  \hspace{0.05\linewidth} 
  \begin{minipage}[b]{0.45\linewidth}
    \centering
    \includegraphics[width=\linewidth]{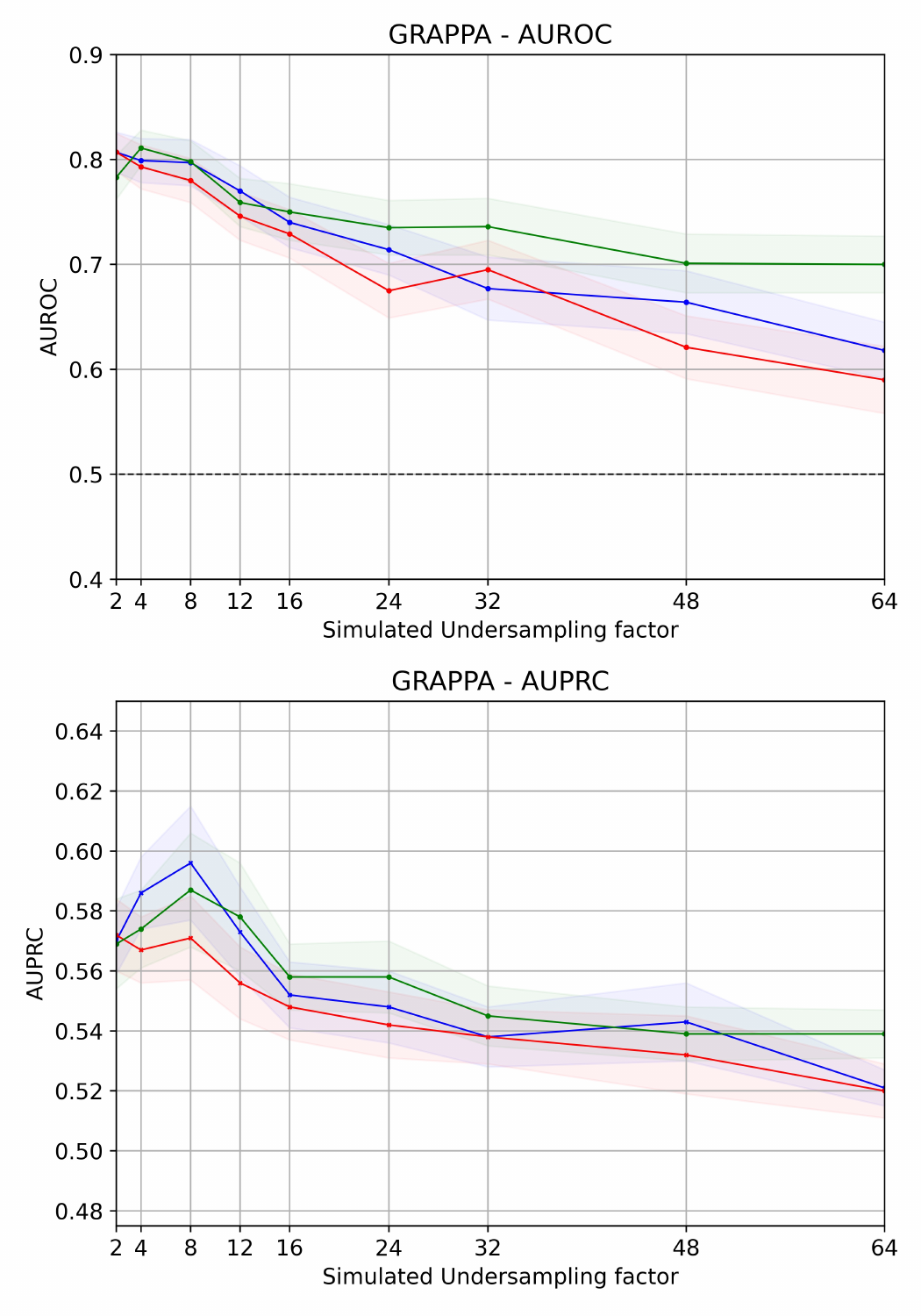}
    \label{fig:results_grappa}
  \end{minipage}
  \caption{PCA (left) and GRAPPA (right) results (top: AUROC; bottom: AUPRC) for the PCA test dataset. Compared are the inputs magnitude in image domain (i) only (blue), magnitude and phase in image domain (red), as well as magnitude in image domain, real and imaginary part in k-Space (k) (green). The AUPRC baseline is 0.051. The model has been trained with the undersampling augmentation of a factor up to x8.}
  \label{fig:results_kspace}
\end{figure*}

\begin{table}[!htb]
\caption{Results for the PCA (left) and GRAPPA (right) test dataset. Compared are the inputs magnitude in image domain (i) only, magnitude and phase in image domain, as well as magnitude in image domain, real and imaginary part in k-Space (k). The model has been trained with the undersampling augmentation of a factor up to x8.}
\centering
\def\arraystretch{1.2}
\resizebox{\columnwidth}{!}{%
\begin{tabular}{lllll}
\toprule
& \multicolumn{2}{c}{GRAPPA} & \multicolumn{2}{c}{PCA} \\
& \multicolumn{1}{c}{AUROC ($\%$) $\uparrow$} & \multicolumn{1}{c}{AUPRC ($\%$) $\uparrow$} & \multicolumn{1}{c}{AUROC ($\%$) $\uparrow$} & \multicolumn{1}{c}{AUPRC ($\%$) $\uparrow$} \\
\midrule
x2 Undersampling & & & & \\
\hspace{1cm} Magnitude (i) & 80.7±1.9 & 57.0±1.1 & 81.3±2.2 & 59.3±1.8 \\
\hspace{1cm} + Phase (i) & 80.7±1.8 & 57.2±1.2 & 80.9±2.1 & 57.7±1.4 \\
\hspace{1cm} + Real \& Imaginary (k) & 78.3±2.1 & 56.9±1.5 & 81.1±1.9 & 57.7±1.4 \\
Simulated x4 Undersampling & & & & \\
\hspace{1cm} Magnitude (i) & 79.9±2.1 & 58.6±1.2 & 80.0±2.3 & 59.4±1.9 \\
\hspace{1cm} + Phase (i) & 79.3±2.1 & 56.7±1.1 & 79.6±2.1 & 57.7±1.4 \\
\hspace{1cm} + Real \& Imaginary (k) & \textbf{81.1±1.7} & 57.4±1.3 & 79.4±2.3 & 57.3±1.4 \\
Simulated x8 Undersampling & & & & \\
\hspace{1cm} Magnitude (i) & 79.7±2.2 & 59.6±1.9 & 78.6±2.6 & 58.5±1.8 \\
\hspace{1cm} + Phase (i) & 78.0±2.1 & 57.1±1.4 & 75.6±2.7 & 56.0±1.5 \\
\hspace{1cm} + Real \& Imaginary (k) & \textbf{79.8±2.0} & 58.7±1.9 & 77.9±2.5 & 57.5±1.4 \\
Simulated x16 Undersampling & & & & \\
\hspace{1cm} Magnitude (i) & 74.9±2.4 & 52.2±1.1 & 67.2±3.1 & 54.6±1.4 \\
\hspace{1cm} + Phase (i) & 72.9±2.3 & 54.8±1.1 & 70.4±2.7 & \textbf{56.8±1.7} \\
\hspace{1cm} + Real \& Imaginary (k) & \textbf{75.0±2.7} & 55.8±1.1 & \textbf{73.3±2.7} & \textbf{56.6±1.6} \\
Simulated x24 Undersampling & & & & \\
\hspace{1cm} Magnitude (i) & 71.4±2.4 & 54.8±1.2 & 61.9±3.1 & 52.8±1.1 \\
\hspace{1cm} + Phase (i) & 67.5±2.6 & 54.2±1.1 & 66.7±2.8 & 54.2±1.1 \\
\hspace{1cm} + Real \& Imaginary (k) & \textbf{73.5±2.6} & \textbf{55.8±1.2} & \textbf{71.4±2.9} & \textbf{56.6±1.8} \\
Simulated x32 Undersampling & & & & \\
\hspace{1cm} Magnitude (i) & 67.7±3.0 & 53.8±1.0 & 53.4±3.6 & 51.2±1.0 \\
\hspace{1cm} + Phase (i) & 69.5±2.8 & 53.8±0.9 & 59.8±3.3 & 52.8±1.2 \\
\hspace{1cm} + Real \& Imaginary (k) & \textbf{73.6±2.7} & \textbf{54.5±1.0} & \textbf{65.9±.3.2} & \textbf{53.5±1.1} \\
Simulated x48 Undersampling & & & & \\
\hspace{1cm} Magnitude (i) & 66.4±3.0 & 54.3±1.0 & 46.7±3.4 & 49.9±0.8 \\
\hspace{1cm} + Phase (i) & 62.1±3.0 & 53.2±1.3 & 56.9±3.4 & 52.3±1.2 \\
\hspace{1cm} + Real \& Imaginary (k) & \textbf{70.1±2.8} & \textbf{53.9±0.9} & \textbf{65.2±3.1} & \textbf{55.2±1.6} \\
Simulated x64 Undersampling & & & & \\
\hspace{1cm} Magnitude (i) & 61.8±2.7 & 52.1±0.6 & 42.0±3.0 & 49.2±0.5 \\
\hspace{1cm} + Phase (i) & 59.0±3.2 & 52.0±0.9 & \textbf{56.6±3.3} & 51.6±0.8 \\
\hspace{1cm} + Real \& Imaginary (k) & \textbf{70.0±2.7} & \textbf{53.9±0.8} & 54.2±3.3 & \textbf{52.1±1.4} \\
\bottomrule
\end{tabular}}
\label{tab:1}
\end{table}

To show the benefits of using k-Space information at higher undersampling rates, we conducted numerical experiments by retrospectively removing lines from the raw k-space, comparing the GRAPPA reconstruction with the PCA coil compression directly in the k-Space without any further reconstruction. It has to be mentioned, that higher digital undersampling rates do not take into account the change in contrasts and artefacts which would occur in real EPI. This will be part of the discussion.
Fig. \ref{fig:results_kspace} shows the results for the PCA and GRAPPA method respectively, with either magnitude only in image domain, magnitude and phase in image domain, as well as magnitude in image domain combined with real and imaginary part in the k-Space, respectively. A detailed list of the results can be found in Tab. \ref{tab:1}. 

For both methods, the benefit of using the additional phase or k-Space information can be shown. In the second experiment the additional k-Space information surpasses the traditional approaches at additional undersampling factors of greater than eight. While the PCA model with only magnitude information starts to fall off strongly at this point with an AUROC of only $67.2\%\pm3.1\%$ at an undersampling factor of $\times24$, especially the model with additional k-Space information is still able to deliver sound results with an AUROC of $73.3\%\pm2.7\%$.
On the PCA data, the magnitude-only model falls under the baseline of random guessing at an undersampling factor of around 40, while the model with additional k-Space information still reaches AUROC of $65.9\%\pm3.2\%$. While the model with additional phase information also beats the magnitude model, it does not outperform the k-Space model. At lower undersampling rates, the phase, as well as the k-Space model reach similar results to the magnitude model with AUROC scores around $80\%$ at both the PCA, as well as the GRAPPA dataset. 

The models trained on the PCA data outperform the models trained on fully reconstructed GRAPPA data for each input type, with AUROC scores of $81.3\%\pm2.2\%$ for the magnitude and $81.1\%\pm1.9\%$ for the magnitude combined with k-Space model on the PCA data, while the GRAPPA models only reach AUROC of $80.7\%\pm1.9\%$ and $78.3\%\pm2.1\%$, respectively.

\section{Discussion}

In this paper, we proposed a novel MRI raw data preprocessing pipeline in combination with a state-of-the art classification neural network to utilize the hidden information in the k-Space for better prediction results as well as faster acquisition and reconstruction times, while still achieving meaningful results for prostate cancer likelihood predictions. 
Using additional k-Space information with a dual network structure we were able to improve the predictions for different data combinations, including the "gold-standard", consisting of ADC, Trace50 and Trace1000 maps.
By applying a PCA instead of time intensive reconstruction techniques such as GRAPPA, as well as using high undersampling factors, the scanning time, as well as the time needed for a first initial diagnosis can be greatly reduced by more than 80\%. 
When performing PCa prediction at undersampling factors greater than eight, the additional information in phase and k-Space show improved stability of the network, enabling it to reach sound results, while the model with only magnitudinal data (which is the data radiologists use for their review) starts to fail making meaningful predictions.

Nevertheless, this study comes with limitations, one being the simple form of undersampling used. The undersampling by evenly leaving out k-Space lines can simulate the corresponding artefacts. It can not reproduce the lower signal-to-noise ratio (SNR) due to fewer acquired measurements \cite{virtue2017empirical} or the reduced motion artifacts due to shorter acquisition times. For DWI imaging with EPI, higher undersampling rates may result in a different image contrast due to reduced T2* weighting as well as distortions due to the shorter echo train.

Performance on more complex undersampling techniques for other imaging sequences will be explored in future experiments. Additionally, the digital undersampling on the GRAPPA output does not reflect the actual data we would use when GRAPPA is applied. Undersampling would be applied before performing reconstruction. Reconstruction techniques such as GRAPPA then try to reconstruct the full k-Space by interpolation. This could lead to better results at low undersampling. At higher undersampling, we suspect a faster performance drop-off as seen in the experiments because of introduced noise amplification in the images due to the ill-posed reconstruction problem at high undersampling rates. This is suspect of further investigations. Because of the time intensive nature of GRAPPA we decided in this work to perform the reconstruction once on the original dataset and performing artificial undersampling afterwards. 

The PCA coil compression approach used in this work only uses the first principal component for coil compression. This approach showed good results, even surpassing ESPIRiT coil combination \cite{uecker2014espirit}, in the case of DWI prostate data and using additional PCA components did not lead to better results. This finding can not be generalized and needs to be evaluated for other sequences. 

Another limitation is the prediction only based on the Pi-RADS score, which tends to be less exact than other ratings for PCa review, such as histopathological examination. In combination with the highly imbalanced dataset, this can lead to unstable training. 

Future work sees the implementation of a fully complex neural network, which can utilize the complex nature of the k-Space properly. This would also include special layers used for k-Space data, such as the Spectral Pooling \cite{rippel2015spectral} or Fourier Convolution \cite{chi2020fast}.
Future work also sees the use of mpMRI raw data for detecting PCa.
Additionally we hypothesize, that the results at small undersampling factors are similar for all approaches due to the fact, that the used neural network is pretrained on data in the magnitudinal image domain. Thus the pretraining on k-Space data will be part of coming projects. 

This work is not only supposed to be a study specifically performed on PCa data, but we want to emphasize the broader picture our results are supposed to show: Clinicians and researchers should start to take into account all the available information acquired in an MRI scan, instead of discarding half of the data right after the preprocessing steps. That way we want to pave the way for better diagnostics in the future by exploiting the raw data domain of MRI scans and at one point eliminating the need for most of the time and resource intensive preprocessing steps, which only goal it is to make the data more interpretable for humans.

\section{Conclusion}

By saving valuable raw data information of MRI DWI images, we were able to show the benefits of using additional information hidden in the k-Space, to predict the PCa likelihodod on DWI data. Applying high undersampling rates in combination with additional k-Space data leads to a reduction of time needed for scanning as well as reconstruction by up to 90\% in comparison to no applied undersampling and commonly used reconstruction techniques such as GRAPPA. 
To the best of our knowledge, this is the first work to explore the feasibility of using MRI raw data for a classification tasks such as PCa likelihood estimation.

Machines need different data than humans. We hope that more research with raw MRI data will emerge in the future, which will lead to improved diagnostics and patient comfort.

\section{References}

\printbibliography[heading=none]

\end{document}